\pdfoutput=1

\documentclass[11pt]{article}

\usepackage{acl}

\usepackage{times}
\usepackage{latexsym}

\usepackage[T1]{fontenc}

\usepackage[utf8]{inputenc}

\usepackage{microtype}

\usepackage{inconsolata}

\usepackage{multirow}
\usepackage{amsmath}
\usepackage{amssymb}
\usepackage{booktabs}
\usepackage{caption}
\usepackage{array}
\newcolumntype{L}[1]{>{\raggedright\let\newline\\\arraybackslash\hspace{0pt}}m{#1}}
\newcolumntype{C}[1]{>{\centering\let\newline\\\arraybackslash\hspace{0pt}}m{#1}}
\newcolumntype{R}[1]{>{\raggedleft\let\newline\\\arraybackslash\hspace{0pt}}m{#1}}
\usepackage{graphicx}
\usepackage{subcaption}
\usepackage{courier}
\usepackage[T1]{fontenc}
\allowdisplaybreaks

\makeatletter
\newcommand{\myfnsymbol}[1]{%
  \expandafter\@myfnsymbol\csname c@#1\endcsname
}
\newcommand{\@myfnsymbol}[1]{%
  \ifcase #1
  0
  \or \TextOrMath{\textasteriskcentered}{*}
  \or 1
  \or 2
  \or 3
  \or 4
  \or 5
  \or 6
  \or 7
  \fi
}
\newcommand{\equalcontributor}{\@myfnsymbol{1}}
\makeatother

\title{FLEUR: An Explainable Reference-Free Evaluation Metric for Image Captioning Using a Large Multimodal Model}

\author{Yebin Lee\textsuperscript{\equalcontributor}\textsuperscript{$\dagger$}, Imseong Park\textsuperscript{\equalcontributor}\textsuperscript{$\diamondsuit$}, Myungjoo Kang\textsuperscript{$\dagger$}\textsuperscript{$\diamondsuit$}\\
  \textsuperscript{$\dagger$}Interdisciplinary Program in Artificial Intelligence, Seoul National University\\
  \textsuperscript{$\diamondsuit$}Department of Mathematical Sciences, Seoul National University\\
  \texttt{\{ylee1846, parkis, mkang\}@snu.ac.kr}}

\begin{document}
\maketitle

\footnotetext[1]{Equal contribution. Correspondence to: Myungjoo Kang}
\setcounter{footnote}{1}

\begin{abstract}
Most existing image captioning evaluation metrics focus on assigning a single numerical score to a caption by comparing it with reference captions. However, these methods do not provide an explanation for the assigned score. Moreover, reference captions are expensive to acquire. In this paper, we propose FLEUR\footnote{We choose a word in French that means `flower', in line with other French-named evaluation metrics.}, an explainable reference-free metric to introduce explainability into image captioning evaluation metrics. By leveraging a large multimodal model, FLEUR can evaluate the caption against the image without the need for reference captions, and provide the explanation for the assigned score. We introduce \textit{score smoothing} to align as closely as possible with human judgment and to be robust to user-defined \textit{grading criteria}. FLEUR achieves high correlations with human judgment across various image captioning evaluation benchmarks and reaches state-of-the-art results on Flickr8k-CF, COMPOSITE, and Pascal-50S within the domain of reference-free evaluation metrics. Our source code and results are publicly available at: \url{https://github.com/Yebin46/FLEUR}.
\end{abstract}

\begin{figure}[t!]
\centering
\includegraphics[width=\columnwidth]{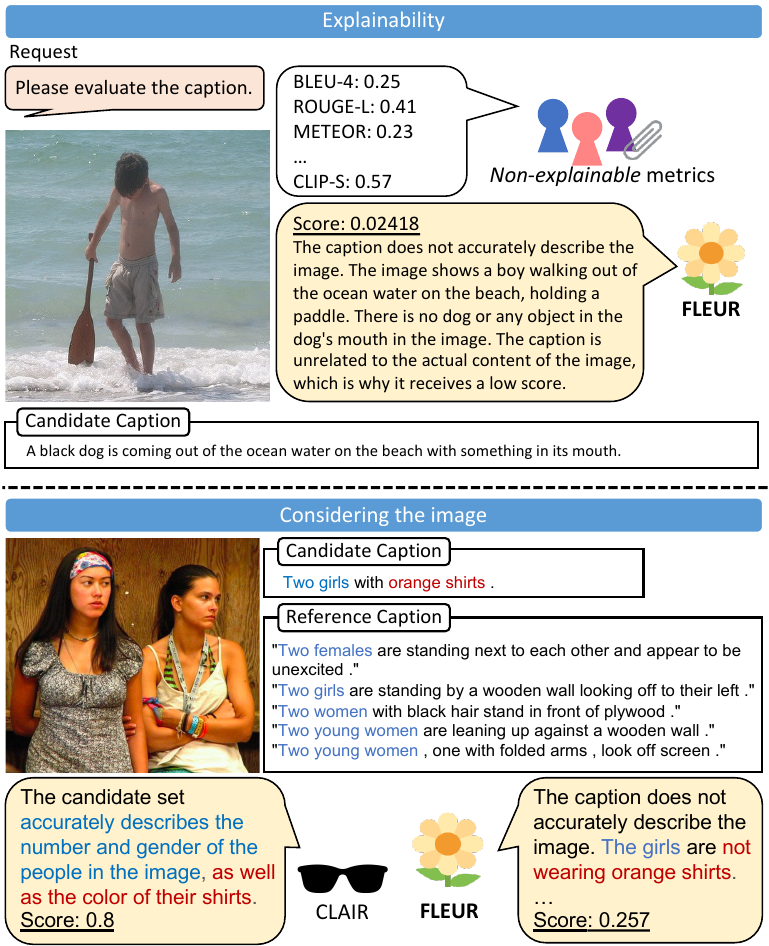} 
\caption{\textbf{Top}: Comparison between other non-explainable metrics and our explainable metric, FLEUR. FLEUR provides the explanation for the assigned score as well. \textbf{Bottom}: Existing explainable metric cannot consider the image. The information highlighted in red in the candidate caption is not present in the reference caption set, causing confusion for that metric.}
\label{fig1}
\end{figure}


\section{Introduction}

Evaluating image captions is essential as it provides a significant indicator of the model’s ability to understand visual and language information effectively (\citealp{mokady2021clipcap}; \citealp{li2023blip}). However, there are two primary challenges with existing image captioning evaluation metrics. Existing methods 1) require reference captions to evaluate candidate captions\footnote{A \textit{reference caption} refers to the human-annotated caption for an image. A \textit{candidate caption} refers to the caption that is to be evaluated.} and 2) lack explainability.

First, traditional image captioning evaluation metrics (\citealp{papineni2002bleuu}; \citealp{lin2004rouge}; \citealp{banerjee2005meteor}) have the drawback of requiring reference captions. These metrics assign scores to candidate captions by comparing them to reference captions. However, in practice, obtaining reference captions is challenging because it requires human annotators to create reference captions. Furthermore, evaluating captions only based on text without direct image comparison cannot yield accurate scores. Therefore, new methods (\citealp{hessel2021clipscore}; \citealp{Sarto_2023_CVPR}; \citealp{hu-etal-2023-infometic}) have emerged that evaluate captions without the need for reference captions by incorporating images.

Second, existing evaluation metrics still lack explainability. Throughout this paper, we clarify the meaning of an \textit{explainable} metric. As defined in \citet{leiter2022explainable}, we embrace the broad concept of \textit{explainability} for metrics. \textbf{The explainability contains the ability to provide an explanation for the score obtained from the metric.} Existing metrics cannot provide intuitive explanations in sentence form. This makes it difficult to discern whether the score is accurate or not. Hence, we categorize metrics incapable of providing descriptive explanations as \textit{non-explainable} metrics (see the top of Figure \ref{fig1}). 

To overcome these two limitations, we propose a reference-\textbf{F}ree exp\textbf{L}ainable \textbf{E}val\textbf{U}ation met\textbf{R}ic (FLEUR) for image captioning. FLEUR can evaluate captions even in the absence of reference captions and provide explanations for the scores by using a large multimodal model (LMM). We introduce \textit{score smoothing} to calibrate the scores from the LMM more finely and make FLEUR robust to prompts. Additionally, we propose a prompt including \textit{grading criteria} for caption evaluation to align the scores more closely with human judgment. It is noteworthy that \textbf{FLEUR is the only caption evaluation metric both explainable and reference-free.}


FLEUR achieves state-of-the-art results across multiple benchmark datasets among the reference-free evaluation metrics, calculated through correlations with human judgment. Furthermore, we demonstrate FLEUR's explainability by comparing its explanations with those of CLAIR \citep{chan-etal-2023-clair}, a reference-based and explainable evaluation metric. We hypothesize that directly viewing the image enables a more accurate and comprehensive evaluation of a candidate caption as shown at the bottom of Figure \ref{fig1}.

Our contributions are as follows:
\begin{itemize}
    \item We propose FLEUR, an explainable reference-free image captioning evaluation metric. FLEUR achieves the highest correlations with human judgment across various benchmark datasets.
    \item To the best of our knowledge, our work is a pioneering work of using an LMM to evaluate image captions. We improve the rating performance of an LMM by introducing score smoothing and grading criteria.
    \item Through a comparison with the reference-based metric CLAIR, we show that FLEUR generates better explanations because it can consider images.
\end{itemize}


\section{Related Works}

\subsection{Image Captioning Evaluation Metrics}

\textbf{Reference-only metrics} compare only the reference captions with the candidate caption (\citealp{vedantam2015cider}; \citealp{anderson2016spice}; \citealp{zhang2019bertscore}). However, reference captions cannot fully encapsulate the image, and natural language inherently contains ambiguity (\citealp{jiang-etal-2019-tiger}). To address this challenge, \textbf{reference+image metrics} have been proposed (\citealp{jiang-etal-2019-tiger}; \citealp{lee-etal-2020-vilbertscore}; \citealp{inan-etal-2021-cosmic-coherence}; \citealp{wada2024}). These metrics consider the image in conjunction with reference captions. Nonetheless, the issue persists that acquiring a set of reference captions for an image is costly and challenging. Therefore, \textbf{reference-free metrics} have been proposed to evaluate candidate captions even in the absence of reference captions (\citealp{hessel2021clipscore}; \citealp{Sarto_2023_CVPR}; \citealp{hu-etal-2023-infometic}). Reference-free metrics address the difficulty of requiring reference captions by evaluating candidate captions against images. However, there is no explainable metric among reference-free metrics. Bringing the advantage of a reference-free metric, FLEUR addresses this issue by using an LMM to provide explanations of the scores.

\begin{figure*}[t]
\centering
\includegraphics[width=\textwidth]{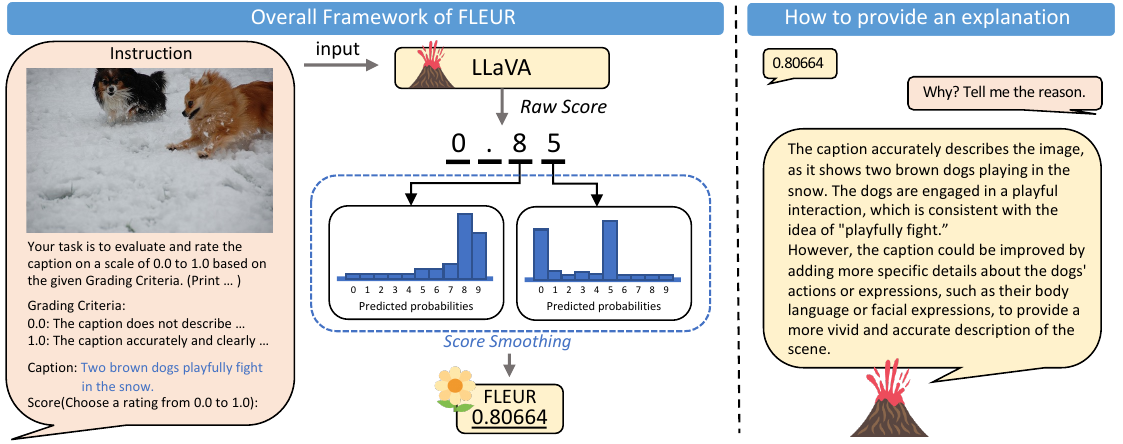} 
\caption{The overall framework of FLEUR. \textbf{Left}: When feeding LLaVA with the prompt containing the grading criteria, image, and the candidate caption for evaluation, FLEUR takes a weighted sum of probabilities of tokens (0 to 9) as the final score. \textbf{Right}: When prompted by the user for the rationale behind the given score, FLEUR provides explanations in a language understandable to humans.}
\label{fig2}
\end{figure*}

\subsection{LLM-based Evaluation Metrics}

\paragraph{Metrics for NLG evaluation} There is a growing interest in leveraging the powerful performance of a large language model (LLM) to evaluate natural language generation (NLG) quality (\citealp{fu2023gptscore}; \citealp{chiang-lee-2023-large}). G-Eval \citep{liu2023geval} assesses the performance of NLG by using an LLM in a form-filling paradigm. \citet{liu2023geval} propose the method that uses the probabilities of output tokens from the LLM, and takes the weighted summation as the final score. G-Eval has to estimate probabilities by sampling 20 times because they use proprietary LLMs like GPT-4 \citep{openai2023gpt4}. In contrast, FLEUR employs an open-source LMM to utilize actual probabilities outputted by the model.

\paragraph{Metrics for caption evaluation} In the field of image caption evaluation, \textbf{CLAIR} \citep{chan-etal-2023-clair} \textbf{first employs text-only LLMs to compare reference caption sets with candidate captions.} CLAIR can provide explanations for the scores, making it an explainable metric. However, \textbf{CLAIR evaluates candidate captions without considering images at all.} If the candidate caption contains information that cannot be inferred from the reference caption set, we cannot expect an accurate score (see the bottom of Figure \ref{fig1}). On the other hand, FLEUR considers the image and therefore provides better evaluation scores and more appropriate explanations.

\section{Method}

The overall framework of FLEUR is depicted in Figure \ref{fig2}. FLEUR derives a \textit{raw score} based on the prompt that accurately reflects \textit{grading criteria} (orange speech bubble on the left of Figure \ref{fig2}), and makes the score continuous by using the probabilities computed by an LMM for \textit{score smoothing} (blue dashed line box in Figure \ref{fig2}). Words marked in \textit{italics} will be explained in detail in the following sections. In this paper, we opt for LLaVA \citep{liu2023improved} as our LMM to introduce score smoothing. Please refer to Appendix \ref{model} for a specification of LLaVA that we used.

\subsection{Prompt for Caption Evaluation}
\label{sec:prompt4capeval}
Our prompt consists of two parts: \textbf{1) \textit{base instruction}, and 2) \textit{grading criteria}.} The base instruction contains essential information for performing caption evaluation, while the grading criteria provide guidelines on how to score during caption evaluation. Our whole prompt is as follows:
\vspace{-0.1cm}
\begin{equation*}
    1)\left\{
    \begin{tabular}{L{6cm}}
        \textit{Your task is to evaluate and rate the caption on a scale of 0.0 to 1.0 based on the given Grading Criteria. (Print Real Number Score ONLY)}
    \end{tabular}
    \right.
\end{equation*}
\begin{equation*}
    2)\left\{
    \begin{tabular}{L{6cm}}
        \textit{Grading Criteria:}\\\\0.0\textit{: The caption does not describe the image at all.}\\1.0\textit{: The caption accurately and clearly describes the image.}
    \end{tabular}
    \right.
\end{equation*}
\begin{quote}
\textit{
\newline
Caption: \{caption\}
\newline
\newline
Score(Choose a rating from 0.0\,to\,1.0):}
\end{quote}

\paragraph{Base instruction} Similar to previous works (\citealp{chan-etal-2023-clair}; \citealp{liu2023geval}), the base instruction explains what task LLaVA should perform. This may include specifications such as the range of scores or requirements for the output format. We request output in the form of a real number between 0 and 1. Additionally, by including a directive to score based on grading criteria, we guide LLaVA to produce convincing scores.

\paragraph{Grading criteria} The grading criteria provide LLaVA with more detailed guidelines on how to evaluate captions based on scores. Intuitively, this approach distinguishes itself from non-explainable metrics by offering the advantage of aligning scores more closely with user intentions. We base the grading criteria for FLEUR on the actual criteria used by human evaluators. Refer to Section \ref{e_gc} for detailed grading criteria selection.

\subsection{Score Smoothing}
\label{score smoothing}
We propose a process called \textit{score smoothing} to calibrate the \textit{raw score}, which is the output of LLaVA. Raw score can be seen as a score before \textit{score smoothing}. Score smoothing utilizes the probabilities of tokens corresponding to digits from 0 to 9. To illustrate, in Figure \ref{fig2}, given an image and the prompt as input, LLaVA outputs ``\texttt{0.85}''. Let's consider the process by which LLaVA generates the output. After LLaVA generates tokens up to the decimal point (i.e., ``\texttt{0.}''), it computes the probabilities of all tokens and chooses the next token ``\texttt{8}'' since token ``\texttt{8}'' has max probability.\footnote{When LLaVA generates the next token, we use greedy decoding to make FLEUR a deterministic metric.} In this computation, we can obtain the probability $p_k$ of the token corresponding to each digit $k$ ($0\le k \le9$) for score smoothing. Similarly, in the process of selecting the token ``\texttt{5}'' for the second decimal place, we can obtain the probability $q_\ell$ of the token corresponding to each digit $\ell$ ($0\le\ell\le9$). Then, the FLEUR score of that image-caption pair is computed as follows:
\begin{equation*}
\begin{aligned}
    &\text{FLEUR} \\&= 0.1\times(0\times p_0+1\times p_1+\cdots+9\times p_9)\\&+0.01\times(0\times q_0+1\times q_1+\cdots+9\times q_9).
\end{aligned}
\end{equation*}

Formally, while generating the raw score, LLaVA computes the probability $p(i,j)$ of each token corresponding to the digit $i$ ($0\le i\le9$) appearing at the $j$-th decimal place ($\,j=1,2$), and we define FLEUR as follows:
\begin{equation*}
\text{FLEUR}=\sum_{j=1}^2 10^{-j}\sum_{i=0}^9i\times p(i,j).
\end{equation*}
Please refer to Appendix \ref{details} for implementation details of FLEUR.

We calibrate the raw score because LLaVA tends to have some scores dominate the distribution of raw scores. This tendency leads to an increase in ties among the scores. In the case of many ties, it is not suitable as an evaluation metric as it cannot discern subtle differences between candidate captions. Even when the raw scores of two image-caption pairs are the same, LLaVA may have different probabilities for those tokens. Consequently, through the application of score smoothing, the scores attain greater granularity, mitigating instances of ties. In Section \ref{e_ss}, we show the positive impact on score smoothing through experiments.

\subsection{RefFLEUR}
In cases where reference captions are provided, FLEUR can be extended to consider the reference caption set when scoring, called \textit{RefFLEUR}. In this case, we provide reference captions and a candidate caption in the prompt. The prompt that overlaps with that of FLEUR is omitted with `...', and the modified parts are indicated with underline. The prompt for RefFLEUR is as follows:

\begin{quote}
    \textit{Your task is to evaluate and rate the \underline{candidate} caption on a scale of 0.0 to 1.0 based on the given Grading Criteria. (Print Real Number Score ONLY)
    \newline
    \dots
    \newline
    1.0: The caption accurately and clearly describes the image.
    \newline
    \newline
    \underline{Reference Captions: \{reference caption set\}}
    \newline
    \newline
    \underline{Candidate}\,Caption: \{candidate caption\}
    \newline
    \dots
    }
\end{quote}

\subsection{Prompt for Explanation}
The explanation for the FLEUR score can be obtained by inputting an additional prompt into LLaVA. As shown on the right of Figure \ref{fig2}, we use the prompt ``\textit{Why? Tell me the reason.}'' Unlike previous studies (\citealp{chan-etal-2023-clair}; \citealp{chiang-lee-2023-large}), we separate the prompts for requesting the score and the explanation to reduce inference time. Examples of the explanations obtained through this prompt are in Figure \ref{fig:clair} and Appendix \ref{examples}.

\renewcommand{\arraystretch}{1.05}
\begin{table*}[t]
\centering
{\small
\begin{tabular}{cclcccccccc}
\toprule
  \multirow{2.5}{*}{\textbf{Type}}& \multirow{2.5}{*}{\textbf{Exp}} & \multirow{2.5}{*}{\textbf{Metric}} & \multicolumn{2}{c}{\textbf{Flickr8k}}  & \textbf{COM} & \multicolumn{5}{c}{\textbf{Pascal-50S} (Accuracy $\uparrow$ )} \\
\cmidrule(lr){4-5}
\cmidrule(lr){6-6}
\cmidrule(lr){7-11}

& && EX ($\tau_c\uparrow$ ) & CF ($\tau_b\uparrow$ ) & ($\tau_c\uparrow$ ) & HC & HI & HM & MM & Avg  \\
\midrule
\multirow{12}{*}{\shortstack{reference\\-based}}&&BLEU-4 & 30.8 & 16.9 & 30.6 & 53.0 & 92.4 & 86.7 & 59.4 & 72.9 \\
&&ROUGE-L & 32.3 & 19.9 & 32.4 & 51.5 & 94.5 & 92.5 & 57.7 & 74.1 \\
&&METEOR & 41.8 & 22.2 & 38.9 & 56.7 & 97.6 & 94.2 & 63.4 & 78.0 \\
&&CIDEr & 43.9 & 24.6 & 37.7 & 53.0 & 98.0 & 91.5 & 64.5 & 76.8 \\
&&SPICE & 44.9 & 24.4 & 40.3 & 52.6 & 93.9 & 83.6 & 48.1 & 69.6 \\
&&BERTScore & 39.2 & 22.8 & 30.1 & 65.4 & 96.2 & 93.3 & 61.4 & 79.1 \\
& $\checkmark$ &CLAIR\footnotemark{} & 48.3 & -- & 61.0 & 52.4 & 99.5 & 89.8 & 73.0 & 78.7 \\
\cmidrule{3-11}
&&TIGEr & 49.3 & -- & 45.4 & 56.0 & \textbf{99.8} & 92.8 & 74.2 & 80.7 \\
&&ViLBERTScore-F & 50.1 & -- & 52.4 & 49.9 & 99.6 & 93.1 & 75.8 & 79.6 \\
&&RefCLIPScore & 53.0 & 36.4 & 55.4 & 64.5 & 99.6 & 95.4 & 72.8 & 83.1 \\
&&RefPAC-S & 55.9 & 37.6 & 57.3 & 67.7 & 99.6 & 96.0 & 75.6 & 84.7 \\
&&Polos & \textbf{56.4} & 37.8 & 57.6 & \textbf{70.0} & 99.6 & 97.4 & \textbf{79.0} & \textbf{86.5} \\
& $\checkmark$ &RefFLEUR (Ours) & 51.9 & \textbf{38.8} & \textbf{64.2} & 68.0 & \textbf{99.8} & \textbf{98.0} & 76.1 & 85.5 \\
\midrule
\multirow{4}{*}{\shortstack{reference\\-free}} &&CLIPScore & 51.2 & 34.4 & 53.8 & 56.5 & 99.3 & 96.4 & 70.4 & 80.7 \\
& &PAC-S & 54.3 & 36.0 & 55.7 & 60.6 & 99.3 & 96.9 & 72.9 & 82.4 \\
& &InfoMetIC+\footnotemark{} & \textbf{55.5} & 36.6 & 59.3 & -- & -- & -- & -- & --  \\
& $\checkmark$&FLEUR (Ours) & 53.0 & \textbf{38.6} & \textbf{63.5} & \textbf{61.3} & \textbf{99.7} & \textbf{97.6} & \textbf{74.2} & \textbf{83.2} \\
\bottomrule
\end{tabular}
}
\caption{
Overall correlation and accuracy comparison with human judgment on Flickr8k-Expert (Flickr8k-EX), Flickr8k-CF, COMPOSITE (COM), and Pascal-50S datasets. Bold indicates the best result in each type. `Exp' stands for `explainable' and checkmarks are applied only to the corresponding metrics. FLEUR is the only metric satisfying both explainable and reference-free. All results except for ours are reported results from prior works.
}
\label{tab:main}
\end{table*}

\section{Experiments}

\subsection{Correlations with Human Judgment}
To assess the quantitative performance of FLEUR, we measure its correlations with human judgment and compare them with other caption evaluation metrics. Following previous studies (\citealp{hessel2021clipscore}; \citealp{hu-etal-2023-infometic}), we evaluate FLEUR on Flickr8k \citep{hodosh2013framing}, COMPOSITE \citep{aditya2015images}, and Pascal-50S \citep{vedantam2015cider} datasets. Please refer to Appendix \ref{dataset} for a comprehensive overview of the datasets.

\paragraph{Performance measures of evaluation metrics} Similar to the previous work \citep{Sarto_2023_CVPR}, we use Kendall’s tau correlation coefficient $\tau$ and accuracy to evaluate the performance of image captioning evaluation metrics. To be more specific, we use tau-c ($\tau_c$) for Flickr8k-Expert and COMPOSITE, and tau-b ($\tau_b$) for Flickr8k-CF. For Pascal-50S, we employ accuracy as the measure, as it assesses whether it aligns with human annotators' preference among two candidate captions.

\footnotetext[5]{We only consider the version of CLAIR that uses GPT-3.5. Due to the disparity in the calculation method of Kendall's correlation coefficient between CLAIR's official results and other metrics, we obtain the CLAIR results publicly available on \url{https://github.com/DavidMChan/clair/issues/2} and calculate Kendall's correlation coefficient directly in the same setting.}
\footnotetext{In the InfoMetIC paper, due to the uncertainty of the Pascal-50S setting, we do not report the score of InfoMetIC. We follow the setting in CLIPScore (\url{https://github.com/jmhessel/clipscore/issues/4}).}

\paragraph{Metrics to compare} Metrics can be divided into two criteria: \textit{whether they are explainable} (`Exp' in Table \ref{tab:main}) and \textit{whether they require reference captions} (`Type' in Table \ref{tab:main}). Excluding our metrics and CLAIR, the remaining metrics are non-explainable because they cannot provide explanations for their scores. Metrics in the first and second block of Table \ref{tab:main} are reference-based metrics. Metrics in the first block only rely on reference captions for evaluation (\textbf{reference-only}), while metrics in the second block evaluate captions along with the image (\textbf{reference+image}). Metrics in the third block are \textbf{reference-free} metrics. Note that \textbf{FLEUR is the only caption evaluation metric both explainable and reference-free.}


\paragraph{Results} Tables \ref{tab:main} presents a comprehensive summary of the results. There are three important takeaways from the results.

\textbf{First}, FLEUR achieves the highest correlation coefficient among reference-free metrics, excluding evaluation on Flickr8k-Expert. Even when including reference-based metrics, FLEUR attains the second-best correlation. Note that RefFLEUR, a variant of FLEUR, achieves the best performance. This implies that FLEUR can evaluate captions closely to human judgment, even in the absence of reference captions.

\textbf{Second}, FLEUR achieves the highest accuracy among reference-free metrics on Pascal-50S. RefFLEUR also achieves the second-best accuracy among reference-based metrics\footnote{Note that Polos is released after the submission of our manuscript. At the time of submission, RefFLEUR had the best performance.}. However, in HC, there is a noticeable difference in accuracy between FLEUR and RefFLEUR. The two candidate captions in HC are both correct human-annotated captions with subtle differences. We speculate that considering reference captions together allows for better detection of these subtle differences, leading to higher accuracy.

\begin{figure*}[t]
\centering
\includegraphics[width=\textwidth]{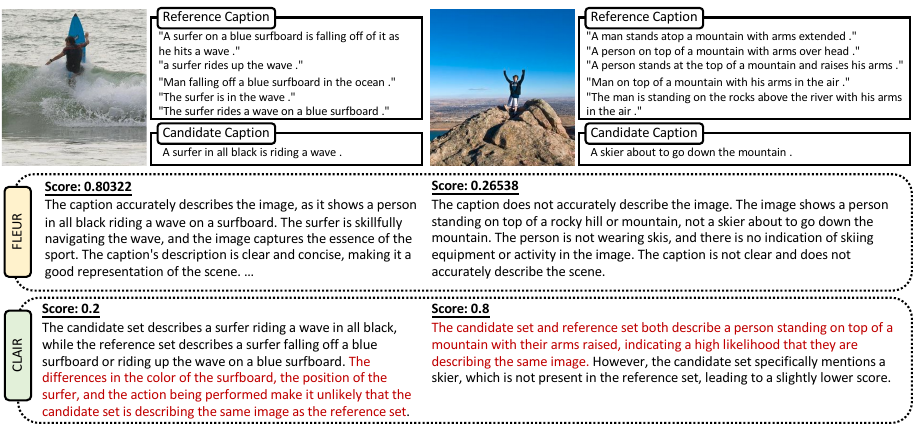} 
\caption{Comparison between the explanation of FLEUR and the explanation of CLAIR. The parts highlighted in red indicate inaccuracies in the explanation. Note that in these examples, FLEUR does not use a reference caption set as input. For spatial reasons, the explanations have been omitted with `...' symbols. The omitted part can be found in Appendix \ref{examples}.}
\label{fig:clair}
\end{figure*}

\textbf{Third}, FLEUR outperforms the explainable reference-only metric CLAIR. We speculate that this is because \textbf{FLEUR can evaluate the caption against the image and generate the more suitable score.} A qualitative comparison among explainable metrics strengthens this speculation (see Section \ref{qualitative}).



\subsection{Comparison with Other Explainable Metric}
\label{qualitative}
We compare the explanations of CLAIR with FLEUR’s explanations for Flickr8k-Expert. As shown in Figure \ref{fig:clair}, CLAIR's explanations often contain inaccuracies. This is particularly evident \textbf{when candidate captions align somewhat with the reference but include additional information not present in the references}. In contrast, FLEUR can evaluate candidate captions based on the image without relying on information from the reference caption.

For example on the left of Figure \ref{fig:clair}, the candidate caption mentions the color of the surfer's clothing (black), but the reference caption set only contains information about the color of the surfboard (blue). CLAIR cannot discern the correctness of such information, and this leads to the generation of incorrect explanations. On the other hand, FLEUR, which can consider the image, can identify the color of the surfer's clothes in the image. Consequently, we speculate that FLEUR provides better evaluation scores and generates more appropriate explanations than CLAIR by directly comparing candidate captions with images.


\renewcommand{\arraystretch}{1.05}
\begin{table}[t]
    \centering
    {
    \small
    \begin{tabular}{lcc}
        \toprule
        \multirow{2.5}{*}{\textbf{Metric}} & \multicolumn{2}{c}{Accuracy ($\uparrow$)}\\
        \cmidrule(lr){2-3}
        & 1-ref & 4-ref\\
        \midrule
        CLAIR & -- & 93.6 \\
        RefPAC-S & 93.7 & 94.9 \\
        RefFLEUR (Ours) & \textbf{97.3} & \textbf{98.4} \\
        \midrule
        PAC-S & 89.9 & 89.9 \\
        FLEUR (Ours) & \textbf{96.8} & \textbf{96.8} \\
        \bottomrule
    \end{tabular}
    }
    \caption{Accuracy comparison on FOIL to evaluate the sensitivity of object hallucination. Note that FLEUR does not need reference captions.}
    \label{tab:foil}
\end{table}

\begin{figure*}[t]
\centering
\includegraphics[width=\textwidth]{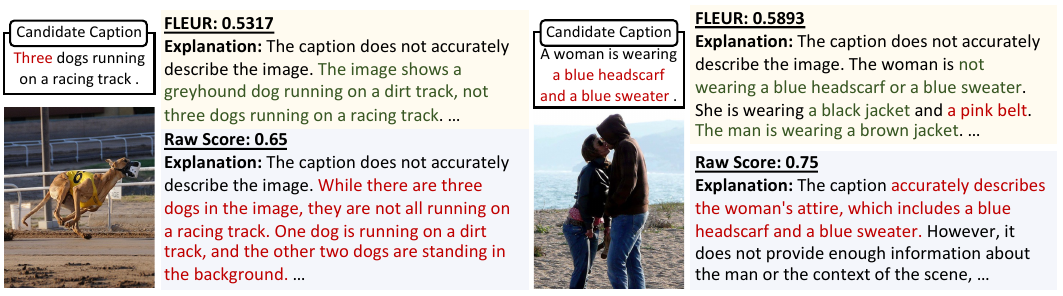} 
\caption{Examples of a FLEUR score and a raw score for the same image-candidate caption pair, along with explanations for each score. 
The parts highlighted in red indicate incorrect captions and incorrect explanations, while the parts marked in green signify correct explanations. 
For spatial reasons, the explanation has been omitted with `...' symbols. The omitted part can be found in Appendix \ref{examples}.}
\label{fig:naive_qual}
\end{figure*}

\subsection{Evaluation of Object Hallucination}
We check whether FLEUR can effectively detect object hallucination, mentioning objects that are not present in images, in FOIL \citep{shekhar-etal-2017-foil} dataset. FOIL is created by replacing a noun phrase related to a single object in the MSCOCO \citep{lin2014microsoft} caption (e.g., changing `cat' to `dog'). We measure the accuracy of detecting object hallucination by comparing the scores of the original caption and the perturbed caption.

In Table \ref{tab:foil}, only the results of the metrics that achieved the highest accuracy on FOIL for each block of Table \ref{tab:main} are presented. FLEUR surpasses the accuracy of the previous state-of-the-art reference-based metric, RefPAC-S. Additionally, FLEUR achieves significantly higher accuracy than the state-of-the-art reference-free metric, PAC-S. Note that all reference-based metrics show a tendency for accuracy to increase with a greater number of reference captions. In contrast, FLEUR achieves overwhelming state-of-the-art performance without using any reference. \textbf{This indicates that FLEUR is highly robust against object hallucination.}

\section{Ablation Studies}

\subsection{Effect of Grading Criteria}
\label{e_gc}
To find the most appropriate grading criteria for evaluating captions, we conduct an ablation study \textit{based on the grading criteria that human annotators actually use} on Flickr8k-Expert. Annotators for Flickr8k-Expert grade candidate captions on a scale of 1 to 4 as integers. The grading criteria referenced by annotators have standards for each score. We normalize integer scores based on our score scale, ranging from 0 to 1, and use them in our grading criteria. The guidelines based on normalized scores are as follows:

\begin{quote}
    \textit{0.0: The caption does not describe the image at all.
    \newline
    0.3: The caption describes minor aspects of the image but does not describe the image.
    \newline
    0.7: The caption almost describes the image with minor mistakes.
    \newline
    1.0: The caption accurately and clearly describes the image.
    }
\end{quote}

\begin{figure}[t]
\centering    
    \begin{subfigure}{0.49\linewidth}
        \centering
        \includegraphics[width=\linewidth]{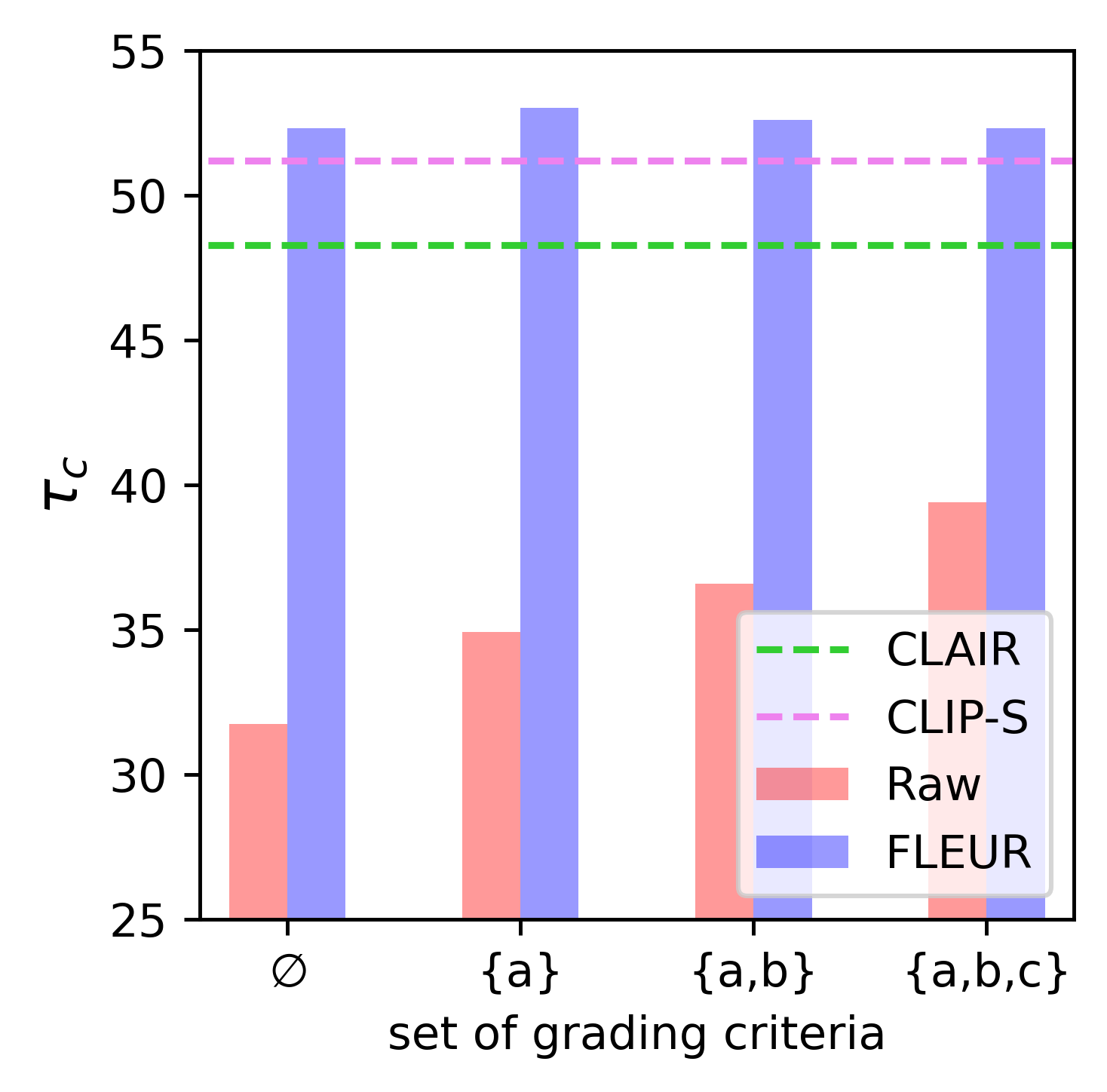} 
        \caption{}
        \label{ab_gc}
    \end{subfigure}
    \begin{subfigure}{0.49\linewidth}
        \centering
        \includegraphics[width=\linewidth]{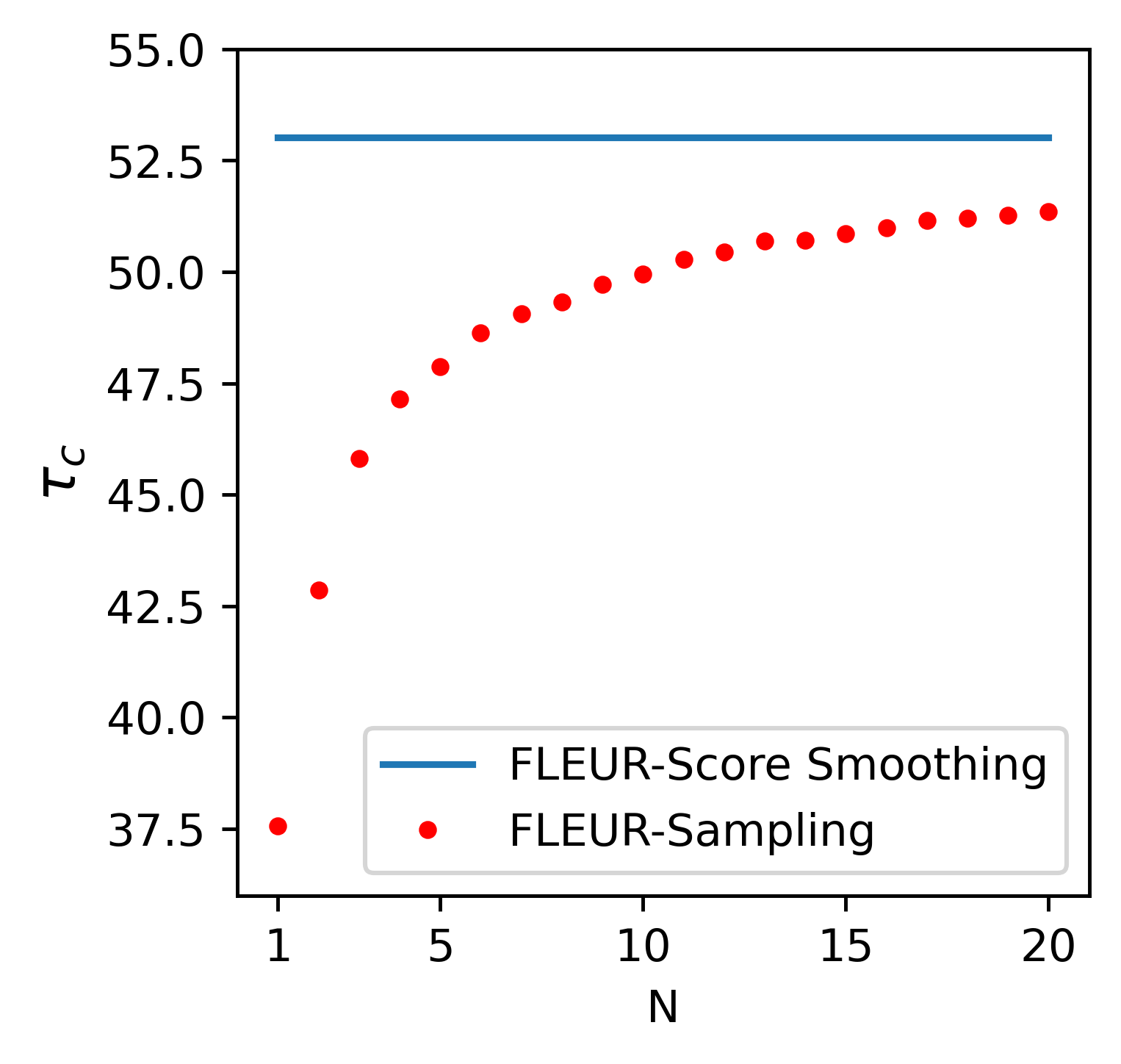} 
        \caption{}
        \label{sampling}
    \end{subfigure}
\caption{(a) Ablation study based on the scores included in the grading criteria. (b) Effect of directly obtaining probabilities.}
\label{fig:ab}
\end{figure}
Figure \ref{ab_gc} is the result of the ablation study based on the scores included in the grading criteria. $\varnothing$ signifies the absence of grading criteria. \texttt{a} represents grading criteria of 0.0 and 1.0, while \texttt{b} and \texttt{c} each denotes grading criteria of 0.3 and 0.7, respectively.\footnote{For instance, \texttt{\{a, b\}} represents grading criteria including guidelines for scores of 0.0, 0.3, and 1.0.} Experimentally, when scores are included in the grading criteria, LLaVA tends to output those scores more frequently. As the number of grading criteria increases, the raw score tends to have a higher correlation with human judgment. Based on this result, we can draw two observations. Introducing detailed grading criteria helps increase correlation with human judgment, and score smoothing makes the model more robust to prompts than raw scores. We select the grading criteria for 0.0 and 1.0, which show the highest correlation with human judgment, as our main method.



\setlength{\tabcolsep}{1.7pt}
\begin{table}[t]
    \centering
    {
    \small
    \begin{tabular}{lcccccccc}
        \toprule
        \multirow{2.2}{*}{\textbf{Metric}} & \multicolumn{2}{c}{\textbf{Flickr8k}}  & \textbf{COM} & \multicolumn{5}{c}{\textbf{Pascal-50S} (Accuracy)}\\
        \cmidrule(lr){2-3}
        \cmidrule(lr){4-4}
        \cmidrule(lr){5-9}
        & EX ($\tau_c$) & CF ($\tau_b$) & ($\tau_c$) & HC & HI & HM & MM & Avg  \\
        \midrule
        Raw & 34.9 & \textbf{51.6} & 58.9 & 26.0 & 99.3 & 92.4 & 42.7 & 65.1 \\
        FLEUR & \textbf{53.0} & 38.6 & \textbf{63.5} & \textbf{61.3} & \textbf{99.7} & \textbf{97.6} & \textbf{74.2} & \textbf{83.2} \\
        \bottomrule
    \end{tabular}
    }
    \caption{Ablation study of score smoothing. We compare the raw score (Raw) and FLEUR performance on benchmark datasets.}
    \label{tab:ab_ss}
\end{table}

\subsection{Effect of Score Smoothing}
\label{e_ss}
The raw score can be considered as an ablated setting that removes score smoothing in FLEUR. We examine the quantitative impact of score smoothing on the performance of the benchmark datasets. As shown in Table \ref{tab:ab_ss}, performance is better with score smoothing (FLEUR) by a large margin for all datasets, except for Flickr8k-CF, compared to the case without score smoothing (Raw).

\setlength{\tabcolsep}{2.5pt}
\begin{table}[t]
    \centering
    {
    \small
    \begin{tabular}{lcccccccc}
        \toprule
        \multirow{2.2}{*}{\textbf{Size}} & \multicolumn{2}{c}{\textbf{Flickr8k}}  & \textbf{COM} & \multicolumn{5}{c}{\textbf{Pascal-50S} (Accuracy)}\\
        \cmidrule(lr){2-3}
        \cmidrule(lr){4-4}
        \cmidrule(lr){5-9}
        & EX ($\tau_c$) & CF ($\tau_b$) & ($\tau_c$) & HC & HI & HM & MM & Avg  \\
        \midrule
        7B & 48.6 & 34.3 & 56.9 & 55.0 & 99.3 & \textbf{97.7} & 71.8 & 81.0 \\
        13B & \textbf{53.0} & \textbf{38.6} & \textbf{63.5} & \textbf{61.3} & \textbf{99.7} & 97.6 & \textbf{74.2} & \textbf{83.2} \\
        \bottomrule
    \end{tabular}
    }
    \caption{Effect of model size of LLaVA.}
    \label{tab:ab_ms}
\end{table}

\subsubsection{Impact on Explanations}

If the scores vary for the same image-caption pair, \textit{would the explanations for the scores also differ?} Surprisingly, when requesting explanations for a FLEUR score and a raw score, the explanations do indeed differ. Therefore, we assess the qualitative impact by comparing explanations of a FLEUR score and a raw score for the same image-caption pair on a per-sample basis.

As shown in Figure \ref{fig:naive_qual}, depending on the score, it can even change whether the candidate caption is considered appropriate or inappropriate in the explanation. In the right side of Figure \ref{fig:naive_qual}, despite the candidate caption inaccurately describing the woman's attire, the raw score assigns a high score (0.75) which contradicts human judgment. The explanation derived from this raw score says that the candidate caption describes the image well. On the other hand, FLEUR assigns a lower score (0.5893) which aligns with human judgment. The explanation derived from this FLEUR score says that the candidate caption does not effectively describe the image. Based on the examples in Figure \ref{fig:naive_qual} and the results in Table \ref{tab:ab_ss}, it can be inferred that FLEUR leads to more accurate explanations. 


\subsubsection{Advantages of Obtaining Probabilities} To assess the impact of directly obtaining probabilities using an open-source model, we compare the performance of sampling to estimate probabilities with directly obtaining probabilities on Flickr8k-Expert. Similar to G-Eval \citep{liu2023geval}, we let some randomness in LLaVA's responses (\texttt{temperature} $\neq 0$), and $N$ samples are drawn to consider the average response as the final score.

Figure \ref{sampling} illustrates the $\tau_{c}$ values for different sampling counts $N$ on Flickr8k-Expert. As $N$ increased, the correlation with human judgment also increased. However, even up to $N=20$, directly obtaining probabilities and applying score smoothing perform better in terms of correlation compared to sampling. \textbf{FLEUR, which uses score smoothing, is deterministic and advantageous from an inference time perspective as it only requires sampling once.}

\subsection{Effect of LMM}
\paragraph{Model size} We compare the impact of LLaVA model size on performance in Table \ref{tab:ab_ms}. In all benchmark datasets excluding HM, the 13B model outperforms the 7B model. In HM, the performance difference between the 7B model and the 13B model is indeed minimal (0.1).

\paragraph{Other models} We assess FLEUR with open-source models, MiniGPT-v2 \citep{chen2023minigptv2}, InstructBLIP \citep{instructblip}, and LLaVA as candidates for our LMM. These models have good performance in various vision-language tasks. However, MiniGPT-v2 outputs scores in the form of ``\texttt{0 5}'' and ``\texttt{0 9}'', instead of the format we expect (i.e., real number). Similarly, on Flickr8k-Expert, InstructBLIP only outputs one of the three scores $\{0.5, 0.8, 1.0\}$, with approximately 99\% of the outputs being 0.5. Despite the poor raw scores of these models, \textbf{the results significantly improved after applying score smoothing} (See Table \ref{tab:ab_lmm}). Since LLaVA still achieves the highest correlation with human judgment, we select LLaVA as our LMM. By finding a prompt tailored to each model, our method can be extended to other LMMs. We leave it to future work.

\setlength{\tabcolsep}{1pt}
\begin{table}[t]
    \centering
    {
    \small
    \begin{tabular}{lcccccccc}
        \toprule
        \multirow{2.2}{*}{\textbf{Model}} & \multicolumn{2}{c}{\textbf{Flickr8k}}  & \textbf{COM} & \multicolumn{5}{c}{\textbf{Pascal-50S} (Accuracy)}\\
        \cmidrule(lr){2-3}
        \cmidrule(lr){4-4}
        \cmidrule(lr){5-9}
        & EX ($\tau_c$) & CF ($\tau_b$) & ($\tau_c$) & HC & HI & HM & MM & Avg  \\
        \midrule
        MiniGPT-v2 & 44.0 & 23.7 & 41.5 & 50.0 & 95.4 & 82.5 & 64.4 & 73.1 \\
        InstructBLIP & 45.6 & 34.0 & 53.5 & 52.2 & 98.8 & 92.9 & 67.5 & 77.9 \\
        FLEUR & \textbf{53.0} & \textbf{38.6} & \textbf{63.5} & \textbf{61.3} & \textbf{99.7} & \textbf{97.6} & \textbf{74.2} & \textbf{83.2} \\
        \bottomrule
    \end{tabular}
    }
    \caption{Ablation study of LMMs. Note that we use slightly different prompts for each model.}
    \label{tab:ab_lmm}
\end{table}

\section{Discussion and Conclusion}
We propose an explainable reference-free metric, FLEUR, achieving notable correlations with human judgment. Our metric utilizes the latest LMM, LLaVA, to compare only an image and a candidate caption without reference captions. Therefore, no additional cost and time such as creating reference captions is needed. Directly comparing an image and a candidate caption allows for appropriate scoring and the generation of a reasonable explanation. This explainability not only unveils the reason behind the evaluated score, but also holds the potential to provide insights that can enhance image captioning model performance, or to be utilized as a dataset by employing score-explanation pairs in training other models.

Note that FLEUR is the first work to use an LMM in caption evaluation metrics. It may be possible to find more suitable prompts through prompt engineering (\citealp{white2023prompt}; \citealp{voronov2024mind}) in order to obtain better scores and explanations. Furthermore, FLEUR can be extended to other LMMs through prompts tailored to the model being used, rather than the prompts we employed. We hope our work can contribute to the explainability of image captioning evaluation metrics.

\section*{Limitations}
Our approach is straightforward and demonstrates good evaluation performance, but as we evaluate image captions using an off-the-shelf LMM, it inherits the issues associated with the LMM itself.

\paragraph{Necessity of post-processing} Despite the request to output only a real number in the prompt, it cannot be guaranteed that the output of the LMM conforms to the request (\citealp{chan-etal-2023-clair}). Rarely, there are outputs other than scores (e.g., \texttt{Score:}). We remove any additional text and extract only the scores. Fortunately, in our experimental setup, there are no instances where scores are not obtained for all benchmark datasets.

\paragraph{Hallucination} There may be concerns about hallucination in FLEUR's explanations. In reality, FLEUR generates nearly accurate explanations when assigning scores similar to human judgment on benchmark datasets. However, when assigning incorrect scores, FLEUR generates explanations with hallucinations (see Appendix \ref{failure} for our failure cases). Nevertheless, this concern is less likely compared to other non-explainable metrics, and as it is actively researched in the current field of LLMs, addressing the inherent issues in LLMs will naturally contribute to resolving this problem.

\paragraph{Preference for LLM-based output} We investigate whether there is a tendency to prefer sentences generated by LLMs, as seen in the previous evaluation metric using LLMs \citep{liu2023geval}. We compare reference captions from Pascal-50S with short descriptions generated by LLaVA, evaluating them using FLEUR. For details on generating short descriptions, please refer to Appendix \ref{short}. As observed in \citet{liu2023geval}, FLEUR assigns an average score 0.1 higher (standard deviation: 0.087) to short descriptions generated by LLaVA compared to reference captions. However, we observe that the captions generated by LLaVA often contain more detailed information than the reference captions. For instance, LLaVA's generated captions include descriptions of the bird's color and species, while the reference captions simply mention `bird'. However, the opposite case also exists, and due to the difficulty in evaluating this issue, it can be considered a limitation of our study.

\paragraph{Inference Time}
Due to the use of an LMM, FLEUR requires longer inference times compared to other metrics. The comparison of inference times is provided in Table \ref{tab:inference_time} in Appendix. To measure inference times, we evaluated metrics on a system equipped with a GeForce RTX 3090 and an Intel Xeon Silver 4210R. While SPICE and CLIPScore have inference times of less than 20 ms per sample, FLEUR and RefFLEUR require approximately 0.70 s and 0.76 s per sample, respectively. However, these times remain within acceptable limits, as evaluating the entire Flickr8k-Expert dataset takes about an hour, which falls within the reasonable range for metrics.

\paragraph{Generation Order of Explanation} Due to the auto-regressive nature of LMMs, generating the explanation after the score (\textit{score-explanation} order) can be considered post-hoc rationalization. To address this, we conduct experiments to generate explanations before scores (\textit{explanation-score} order). As a result, generating in explanation-score order achieves a high human judgment correlation similar to the FLEUR (see Appendix \ref{appendix:post-hoc} for details). This demonstrates score smoothing is relatively robust to prompts, and FLEUR that obtained by conducting caption evaluation in the score-explanation order is meaningful. Despite being considered post-hoc rationalization, the advantage of our approach lies in enhancing understanding of scores compared to existing metrics that do not provide explanations and facilitating factual verification.

\section*{Acknowledgements}
We thank Jaewoong Choi and Jaemoo Choi for their valuable feedback. We thank all the anonymous reviewers for their constructive comments. This work was supported by NRF grant[RS-2024-00421203], and MSIT/IITP[2021-0-01343-004, Artificial Intelligence Graduate School Program (Seoul National University)].



\begin{thebibliography}{35}
\expandafter\ifx\csname natexlab\endcsname\relax\def\natexlab#1{#1}\fi

\bibitem[{Achiam et~al.(2023)Achiam, Adler, Agarwal, Ahmad, Akkaya, Aleman, Almeida, Altenschmidt, Altman, Anadkat et~al.}]{openai2023gpt4}
Josh Achiam, Steven Adler, Sandhini Agarwal, Lama Ahmad, Ilge Akkaya, Florencia~Leoni Aleman, Diogo Almeida, Janko Altenschmidt, Sam Altman, Shyamal Anadkat, et~al. 2023.
\newblock Gpt-4 technical report.
\newblock \emph{arXiv preprint arXiv:2303.08774}.

\bibitem[{Aditya et~al.(2015)Aditya, Yang, Baral, Fermuller, and Aloimonos}]{aditya2015images}
Somak Aditya, Yezhou Yang, Chitta Baral, Cornelia Fermuller, and Yiannis Aloimonos. 2015.
\newblock From images to sentences through scene description graphs using commonsense reasoning and knowledge.
\newblock \emph{arXiv preprint arXiv:1511.03292}.

\bibitem[{Anderson et~al.(2016)Anderson, Fernando, Johnson, and Gould}]{anderson2016spice}
Peter Anderson, Basura Fernando, Mark Johnson, and Stephen Gould. 2016.
\newblock Spice: Semantic propositional image caption evaluation.
\newblock In \emph{ECCV}.

\bibitem[{Banerjee and Lavie(2005)}]{banerjee2005meteor}
Satanjeev Banerjee and Alon Lavie. 2005.
\newblock {METEOR}: An automatic metric for {MT} evaluation with improved correlation with human judgments.
\newblock In \emph{ACL Workshop on Intrinsic and Extrinsic Evaluation Measures for Machine Translation and/or Summarization}.

\bibitem[{Chan et~al.(2023)Chan, Petryk, Gonzalez, Darrell, and Canny}]{chan-etal-2023-clair}
David Chan, Suzanne Petryk, Joseph Gonzalez, Trevor Darrell, and John Canny. 2023.
\newblock {CLAIR}: Evaluating image captions with large language models.
\newblock In \emph{EMNLP}.

\bibitem[{Chen et~al.(2023)Chen, Zhu, Shen, Li, Liu, Zhang, Krishnamoorthi, Chandra, Xiong, and Elhoseiny}]{chen2023minigptv2}
Jun Chen, Deyao Zhu, Xiaoqian Shen, Xiang Li, Zechu Liu, Pengchuan Zhang, Raghuraman Krishnamoorthi, Vikas Chandra, Yunyang Xiong, and Mohamed Elhoseiny. 2023.
\newblock Minigpt-v2: large language model as a unified interface for vision-language multi-task learning.
\newblock \emph{arXiv preprint arXiv:2310.09478}.

\bibitem[{Chiang and Lee(2023)}]{chiang-lee-2023-large}
Cheng-Han Chiang and Hung-yi Lee. 2023.
\newblock Can large language models be an alternative to human evaluations?
\newblock In \emph{ACL}.

\bibitem[{Chiang et~al.(2023)Chiang, Li, Lin, Sheng, Wu, Zhang, Zheng, Zhuang, Zhuang, Gonzalez, Stoica, and Xing}]{vicuna2023}
Wei-Lin Chiang, Zhuohan Li, Zi~Lin, Ying Sheng, Zhanghao Wu, Hao Zhang, Lianmin Zheng, Siyuan Zhuang, Yonghao Zhuang, Joseph~E. Gonzalez, Ion Stoica, and Eric~P. Xing. 2023.
\newblock Vicuna: An open-source chatbot impressing gpt-4 with 90\%* chatgpt quality.
\newblock \url{https://lmsys.org/blog/2023-03-30-vicuna/}.
\newblock Accessed: 2023-11-29.

\bibitem[{Dai et~al.(2023)Dai, Li, Li, Tiong, Zhao, Wang, Li, Fung, and Hoi}]{instructblip}
Wenliang Dai, Junnan Li, Dongxu Li, Anthony Meng~Huat Tiong, Junqi Zhao, Weisheng Wang, Boyang Li, Pascale Fung, and Steven Hoi. 2023.
\newblock Instructblip: Towards general-purpose vision-language models with instruction tuning.
\newblock \emph{arXiv preprint arXiv:2305.06500}.

\bibitem[{Fu et~al.(2023)Fu, Ng, Jiang, and Liu}]{fu2023gptscore}
Jinlan Fu, See-Kiong Ng, Zhengbao Jiang, and Pengfei Liu. 2023.
\newblock Gptscore: Evaluate as you desire.
\newblock \emph{arXiv preprint arXiv:2302.04166}.

\bibitem[{Hessel et~al.(2021)Hessel, Holtzman, Forbes, Le~Bras, and Choi}]{hessel2021clipscore}
Jack Hessel, Ari Holtzman, Maxwell Forbes, Ronan Le~Bras, and Yejin Choi. 2021.
\newblock {CLIPS}core: A reference-free evaluation metric for image captioning.
\newblock In \emph{EMNLP}.

\bibitem[{Hodosh et~al.(2013)Hodosh, Young, and Hockenmaier}]{hodosh2013framing}
Micah Hodosh, Peter Young, and Julia Hockenmaier. 2013.
\newblock Framing image description as a ranking task: Data, models and evaluation metrics.
\newblock \emph{JAIR}.

\bibitem[{Hu et~al.(2023)Hu, Chen, Zhang, and Jin}]{hu-etal-2023-infometic}
Anwen Hu, Shizhe Chen, Liang Zhang, and Qin Jin. 2023.
\newblock {I}nfo{M}et{IC}: An informative metric for reference-free image caption evaluation.
\newblock In \emph{ACL}.

\bibitem[{Inan et~al.(2021)Inan, Sharma, Khalid, Soricut, Stone, and Alikhani}]{inan-etal-2021-cosmic-coherence}
Mert Inan, Piyush Sharma, Baber Khalid, Radu Soricut, Matthew Stone, and Malihe Alikhani. 2021.
\newblock {COSM}ic: A coherence-aware generation metric for image descriptions.
\newblock In \emph{EMNLP Findings}.

\bibitem[{Jiang et~al.(2019)Jiang, Huang, Zhang, Wang, Zhang, Gan, Diesner, and Gao}]{jiang-etal-2019-tiger}
Ming Jiang, Qiuyuan Huang, Lei Zhang, Xin Wang, Pengchuan Zhang, Zhe Gan, Jana Diesner, and Jianfeng Gao. 2019.
\newblock {TIGE}r: Text-to-image grounding for image caption evaluation.
\newblock In \emph{EMNLP}.

\bibitem[{Lee et~al.(2020)Lee, Yoon, Dernoncourt, Kim, Bui, and Jung}]{lee-etal-2020-vilbertscore}
Hwanhee Lee, Seunghyun Yoon, Franck Dernoncourt, Doo~Soon Kim, Trung Bui, and Kyomin Jung. 2020.
\newblock {V}i{LBERTS}core: Evaluating image caption using vision-and-language {BERT}.
\newblock In \emph{First Workshop on Evaluation and Comparison of NLP Systems}.

\bibitem[{Leiter et~al.(2022)Leiter, Lertvittayakumjorn, Fomicheva, Zhao, Gao, and Eger}]{leiter2022explainable}
Christoph Leiter, Piyawat Lertvittayakumjorn, Marina Fomicheva, Wei Zhao, Yang Gao, and Steffen Eger. 2022.
\newblock Towards explainable evaluation metrics for natural language generation.
\newblock \emph{arXiv preprint arXiv:2203.11131}.

\bibitem[{Li et~al.(2023)Li, Li, Savarese, and Hoi}]{li2023blip}
Junnan Li, Dongxu Li, Silvio Savarese, and Steven Hoi. 2023.
\newblock Blip-2: Bootstrapping language-image pre-training with frozen image encoders and large language models.
\newblock In \emph{ICML}.

\bibitem[{Lin(2004)}]{lin2004rouge}
Chin-Yew Lin. 2004.
\newblock {ROUGE}: A package for automatic evaluation of summaries.
\newblock In \emph{Text Summarization Branches Out}.

\bibitem[{Lin et~al.(2014)Lin, Maire, Belongie, Hays, Perona, Ramanan, Doll{\'a}r, and Zitnick}]{lin2014microsoft}
Tsung-Yi Lin, Michael Maire, Serge Belongie, James Hays, Pietro Perona, Deva Ramanan, Piotr Doll{\'a}r, and C.~Lawrence Zitnick. 2014.
\newblock Microsoft coco: Common objects in context.
\newblock In \emph{ECCV}.

\bibitem[{Liu et~al.(2023{\natexlab{a}})Liu, Li, Li, and Lee}]{liu2023improved}
Haotian Liu, Chunyuan Li, Yuheng Li, and Yong~Jae Lee. 2023{\natexlab{a}}.
\newblock Improved baselines with visual instruction tuning.
\newblock \emph{arXiv preprint arXiv:2310.03744}.

\bibitem[{Liu et~al.(2023{\natexlab{b}})Liu, Li, Wu, and Lee}]{liu2023visual}
Haotian Liu, Chunyuan Li, Qingyang Wu, and Yong~Jae Lee. 2023{\natexlab{b}}.
\newblock Visual instruction tuning.
\newblock \emph{arXiv preprint arXiv:2304.08485}.

\bibitem[{Liu et~al.(2023{\natexlab{c}})Liu, Iter, Xu, Wang, Xu, and Zhu}]{liu2023geval}
Yang Liu, Dan Iter, Yichong Xu, Shuohang Wang, Ruochen Xu, and Chenguang Zhu. 2023{\natexlab{c}}.
\newblock G-eval: Nlg evaluation using gpt-4 with better human alignment.
\newblock \emph{arXiv preprint arXiv:2303.16634}.

\bibitem[{Mokady et~al.(2021)Mokady, Hertz, and Bermano}]{mokady2021clipcap}
Ron Mokady, Amir Hertz, and Amit~H Bermano. 2021.
\newblock Clipcap: Clip prefix for image captioning.
\newblock \emph{arXiv preprint arXiv:2111.09734}.

\bibitem[{Papineni et~al.(2002)Papineni, Roukos, Ward, and Zhu}]{papineni2002bleuu}
Kishore Papineni, Salim Roukos, Todd Ward, and Wei-Jing Zhu. 2002.
\newblock {B}leu: a method for automatic evaluation of machine translation.
\newblock In \emph{ACL}.

\bibitem[{Radford et~al.(2021)Radford, Kim, Hallacy, Ramesh, Goh, Agarwal, Sastry, Askell, Mishkin, Clark, Krueger, and Sutskever}]{radford2021learning}
Alec Radford, Jong~Wook Kim, Chris Hallacy, Aditya Ramesh, Gabriel Goh, Sandhini Agarwal, Girish Sastry, Amanda Askell, Pamela Mishkin, Jack Clark, Gretchen Krueger, and Ilya Sutskever. 2021.
\newblock Learning transferable visual models from natural language supervision.
\newblock In \emph{ICML}.

\bibitem[{Sarto et~al.(2023)Sarto, Barraco, Cornia, Baraldi, and Cucchiara}]{Sarto_2023_CVPR}
Sara Sarto, Manuele Barraco, Marcella Cornia, Lorenzo Baraldi, and Rita Cucchiara. 2023.
\newblock Positive-augmented contrastive learning for image and video captioning evaluation.
\newblock In \emph{CVPR}.

\bibitem[{Shekhar et~al.(2017)Shekhar, Pezzelle, Klimovich, Herbelot, Nabi, Sangineto, and Bernardi}]{shekhar-etal-2017-foil}
Ravi Shekhar, Sandro Pezzelle, Yauhen Klimovich, Aur{\'e}lie Herbelot, Moin Nabi, Enver Sangineto, and Raffaella Bernardi. 2017.
\newblock {FOIL} it! find one mismatch between image and language caption.
\newblock In \emph{ACL}.

\bibitem[{Touvron et~al.(2023)Touvron, Lavril, Izacard, Martinet, Lachaux, Lacroix, Rozi{\`e}re, Goyal, Hambro, Azhar et~al.}]{touvron2023llama}
Hugo Touvron, Thibaut Lavril, Gautier Izacard, Xavier Martinet, Marie-Anne Lachaux, Timoth{\'e}e Lacroix, Baptiste Rozi{\`e}re, Naman Goyal, Eric Hambro, Faisal Azhar, et~al. 2023.
\newblock Llama: Open and efficient foundation language models.
\newblock \emph{arXiv preprint arXiv:2302.13971}.

\bibitem[{Vedantam et~al.(2015)Vedantam, Zitnick, and Parikh}]{vedantam2015cider}
R.~Vedantam, C.~Zitnick, and D.~Parikh. 2015.
\newblock Cider: Consensus-based image description evaluation.
\newblock In \emph{CVPR}.

\bibitem[{Voronov et~al.(2024)Voronov, Wolf, and Ryabinin}]{voronov2024mind}
Anton Voronov, Lena Wolf, and Max Ryabinin. 2024.
\newblock Mind your format: Towards consistent evaluation of in-context learning improvements.
\newblock \emph{arXiv preprint arXiv:2401.06766}.

\bibitem[{Wada et~al.(2024)Wada, Kaneda, Saito, and Sugiura}]{wada2024}
Yuiga Wada, Kanta Kaneda, Daichi Saito, and Komei Sugiura. 2024.
\newblock {Polos: Multimodal Metric Learning from Human Feedback for Image Captioning}.
\newblock In \emph{CVPR}.

\bibitem[{White et~al.(2023)White, Fu, Hays, Sandborn, Olea, Gilbert, Elnashar, Spencer-Smith, and Schmidt}]{white2023prompt}
Jules White, Quchen Fu, Sam Hays, Michael Sandborn, Carlos Olea, Henry Gilbert, Ashraf Elnashar, Jesse Spencer-Smith, and Douglas~C Schmidt. 2023.
\newblock A prompt pattern catalog to enhance prompt engineering with chatgpt.
\newblock \emph{arXiv preprint arXiv:2302.11382}.

\bibitem[{Young et~al.(2014)Young, Lai, Hodosh, and Hockenmaier}]{young-etal-2014-image}
Peter Young, Alice Lai, Micah Hodosh, and Julia Hockenmaier. 2014.
\newblock From image descriptions to visual denotations: New similarity metrics for semantic inference over event descriptions.
\newblock \emph{TACL}.

\bibitem[{Zhang et~al.(2020)Zhang, Kishore, Wu, Weinberger, and Artzi}]{zhang2019bertscore}
Tianyi Zhang, Varsha Kishore, Felix Wu, Kilian~Q. Weinberger, and Yoav Artzi. 2020.
\newblock Bertscore: Evaluating text generation with bert.
\newblock In \emph{ICLR}.

\end{thebibliography}

\appendix


\section{Model Details}
\label{model}
LLaVA \citep{liu2023visual} is a multimodal large language model that integrates a pretrained CLIP \citep{radford2021learning} ViT-L visual encoder and a large language model (LLM) using MLP projection. The model was trained using MSCOCO \citep{lin2014microsoft} bounding box and caption datasets, with instructions generated by the text-only GPT-4 \citep{openai2023gpt4}. Despite being fully open source, LLaVA demonstrated performance comparable to GPT-3.5 across various downstream tasks.

The version of the model used in our study is `\texttt{liuhaotian/llava-v1.5-13b}'. The differences between LLaVA v1.0 \citep{liu2023visual} and LLaVA v1.5 \citep{liu2023improved} lie in the types of base models, the amount of data used for training, and the addition of a simple formatting prompt. For the visual encoder, LLaVA v1.0 utilizes CLIP ViT-L/14, while LLaVA v1.5 uses CLIP ViT-L/14@336px. As for the LLM, LLaVA v1.0 employs LLaMA \citep{touvron2023llama}, and LLaVA v1.5 uses Vicuna \citep{vicuna2023}.

\section{Implementation Details}
\label{details}
The following is an example of FLEUR calculation for the sample in Figure \ref{fig2}:
\begin{align*}
    0.80664&\\
    \approx0.1&\times\big(0\times0.003021240234375\\
    &+1\times0.00128936767578125\\
    &+2\times0.0018758773803710938\\
    &+3\times0.00353240966796875\\
    &+4\times0.00827789306640625\\
    &+5\times0.03350830078125\\
    &+6\times0.07672119140625\\
    &+7\times0.2117919921875\\
    &+8\times0.383544921875\\
    &+9\times0.2763671875\big)\\
    +0.01&\times\big(0\times0.0450439453125\\
    &+1\times0.035614013671875\\
    &+2\times0.050628662109375\\
    &+3\times0.044342041015625\\
    &+4\times0.0400390625\\
    &+5\times0.3515625\\
    &+6\times0.048309326171875\\
    &+7\times0.041961669921875\\
    &+8\times0.04681396484375\\
    &+9\times0.035888671875\big).
\end{align*}

Experimentally though not very common, score smoothing poses a slight issue when LLaVA outputs ``\texttt{1.0}''. This occurs because only tokens after the decimal point are considered, disregarding the units place. To address this, we assume that the probability of obtaining ``\texttt{0}'' in the units place corresponds to the likelihood of LLaVA outputting a raw score of 0.9. Consequently, the final score is determined by the sum of the probability of obtaining ``\texttt{0}'' in the units place multiplied by 0.9, and the probability of obtaining ``\texttt{1}'' in the units place.

\section{Datasets}
\label{dataset}
Flickr8k-Expert \citep{hodosh2013framing} consists of 5,664 image-caption pairs with 1,000 images, each image having 4 or 5 corresponding reference captions. Three expert annotators evaluate each image-caption pair on a scale from 1 (not a match) to 4 (perfect match).

Flickr8k-CF \citep{hodosh2013framing} comprises 47,830 image-caption pairs with 1,000 images, each having approximately 5 reference captions. An evaluation by three annotators involves determining whether the image-caption pair is a match, categorized as either ``yes'' or ``no''. The score for each pair is the proportion of ``yes'' responses.

COMPOSITE \citep{aditya2015images} contains 3,995 images sourced from Flickr8k (997 images), Flickr30k (991 images; \citealp{young-etal-2014-image}), and MSCOCO (2,007 images; \citealp{lin2014microsoft}). Each image is associated with 3 candidate captions and approximately 5 reference captions. Image-caption pairs in COMPOSITE are scored on a scale from 1 (not a match) to 5 (perfect match).

Pascal-50S \citep{vedantam2015cider} includes 4,000 caption pairs with 1,000 images, and a label indicating which of the two captions is deemed correct by 48 annotators. Approximately 50 reference captions are linked to each image. The caption pairs in Pascal-50S are categorized based on the composition of the two captions: HC denotes two correct human-written captions; HI indicates two human-written captions with one correct and one incorrect; HM represents one human-written caption and one machine-generated caption; MM signifies two machine-generated captions.

FOIL \citep{shekhar-etal-2017-foil} is created from 32,150 images extracted from MSCOCO. It consists of 99,480 image-caption pairs, where the captions are made as follows: one from the MSCOCO captions of the image, and the other by replacing one noun in the selected caption with an incorrect but similar word.

\section{More Examples of Explanations}
\label{examples}
\begin{figure*}[t]
\centering
\includegraphics[width=\textwidth]{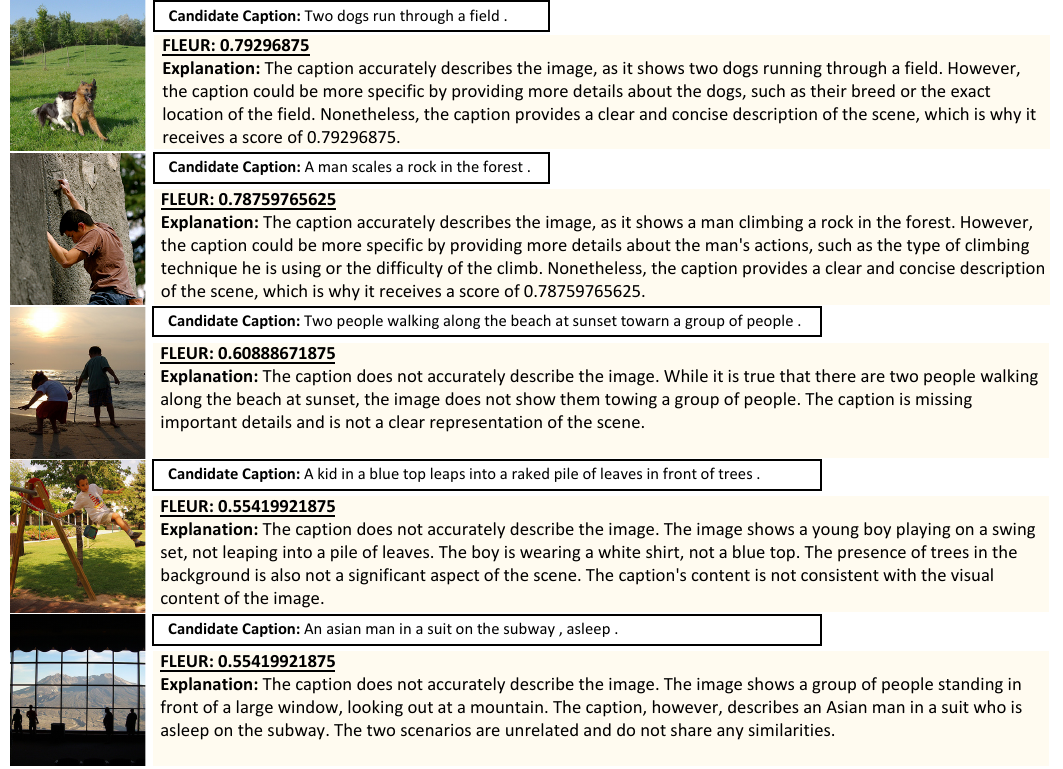} 
\caption{More examples of FLEUR.}
\label{fig:examples}
\end{figure*}

Figure \ref{fig:examples} provides additional examples of explanations generated by LLaVA for FLEUR. Table \ref{omitted} displays the complete explanations that are omitted in the main paper's figures, indicated by `...' notation. Table \ref{fig:c_examples} shows more examples of explanations for FLEUR and explanations for CLAIR.

\renewcommand{\arraystretch}{1.5}
\setlength{\tabcolsep}{1.7pt}
\begin{table}[t]
    \centering
    {
    \small
    \begin{tabular}{lr}
        \toprule
        \textbf{Metric} & Inference time (sec)\\
        \midrule
        SPICE & 0.0177 \\
        CLIPScore & 0.0017 \\
        RefCLIPScore & 0.0028 \\
        FLEUR & 0.70 \\
        RefFLEUR & 0.76 \\
        \bottomrule
    \end{tabular}
    }
    \caption{Comparison of inference times}
    \label{tab:inference_time}
\end{table}

\section{Failure Cases}
\label{failure}
\begin{figure*}[t]
\centering
\includegraphics[width=\textwidth]{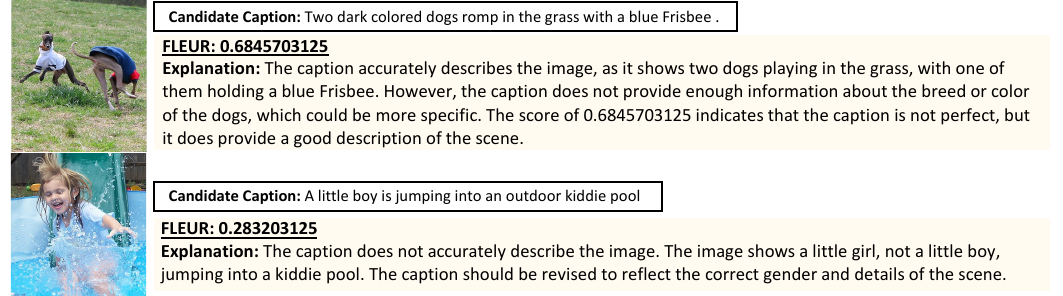} 
\caption{Examples for which explanations of FLEUR failed.}
\label{fig:failure}
\end{figure*}

Figure \ref{fig:failure} illustrates cases where the explanations generated by LLaVA for FLEUR are inconsistent with reality. Analyzing instances in Flickr8k-Expert where FLEUR's judgment does not align with human judgment, two types of errors can be identified. First, samples where LLaVA fails to accurately comprehend the image, as seen in the top of Figure \ref{fig:failure} where it appears to have mistaken a dog's blue clothing for a frisbee. Second, samples where the model exhibits bias, as evidenced in the bottom of Figure \ref{fig:failure} where LLaVA incorrectly identifies a boy with long hair as a girl. This is speculated to be due to biases introduced during the pretraining of LLaVA.

\setlength{\tabcolsep}{5pt} 
\renewcommand{\arraystretch}{1.5}
\begin{table*}[t]
\centering
{\small
\begin{tabular}{llp{5cm}p{7cm}}
\toprule
         & Metric & Candidate caption & Whole explanation\\
         \midrule
Figure \ref{fig:clair} & FLEUR  & A surfer in all black is riding a wave .                 & The caption accurately describes the image, as it shows a person in all black riding a wave on a surfboard. The surfer is skillfully navigating the wave, and the image captures the essence of the sport. The caption's description is clear and concise, making it a good representation of the scene. However, it could be slightly improved by adding more details about the surfer's stance or the type of surfboard being used. \\
Figure \ref{fig:naive_qual} & FLEUR  & Three dogs running on a racing track .                   & The caption does not accurately describe the image. The image shows a greyhound dog running on a dirt track, not three dogs running on a racing track. The caption is missing important details, such as the fact that the dog is wearing a numbered vest, which is a common practice in dog racing. The caption could be improved by providing more specific details about the dog and its actions, as well as the context of the scene. \\
Figure \ref{fig:naive_qual} & Raw  & Three dogs running on a racing track .                   & The caption does not accurately describe the image. While there are three dogs in the image, they are not all running on a racing track. One dog is running on a dirt track, and the other two dogs are standing in the background. The caption should be more specific about the actions of the dogs and their surroundings. \\
Figure \ref{fig:naive_qual} & FLEUR  & A woman is wearing a blue headscarf and a blue sweater . & The caption does not accurately describe the image. The woman is not wearing a blue headscarf or a blue sweater. She is wearing a black jacket and a pink belt. The man is wearing a brown jacket. The image shows a couple kissing on a beach, but the caption does not provide enough information to accurately describe the scene. \\
Figure \ref{fig:naive_qual} & Raw  & A woman is wearing a blue headscarf and a blue sweater . & The caption accurately describes the woman's attire, which includes a blue headscarf and a blue sweater. However, it does not provide enough information about the man or the context of the scene, such as the beach setting or the couple's relationship. The caption could be improved by adding more details about the man and the environment to provide a more complete description of the image. \\
\bottomrule
\end{tabular}
}
\caption{Full sentences of omitted parts of explanations in the main figures.}
\label{omitted}
\end{table*}

\section{Short Description Generation Details}
\label{short}
We create short descriptions for images in Pascal-50S using LLaVA and compare the score difference with reference captions. The prompt used to obtain short descriptions is ``\textit{Please give me a short caption of the image.}'' Given LLaVA's tendency to generate detailed descriptions, we explicitly instruct it to create short captions. When generating captions, we use LLaVA's hyperparameters with \texttt{top\_p} set to 0.7 and \texttt{temperature} set to 0.2.

\section{Generation Order between Scores and Explanations}
\label{appendix:post-hoc}
Using the Flickr8k-Expert dataset, we modify the order of prompts to generate explanations first, followed by scores. Except for modifying the base instructions to generate explanations before scores, we use the same grading criteria as proposed in FLEUR and apply score smoothing. As a result, generating in explanation-score order achieve $\tau_c=50.3$ that do not differ significantly from FLEUR's result $\tau_c=53.0$.

\begin{table*}[t]
\centering
{\small
\begin{tabular}{lp{14cm}}
\toprule
\textbf{Metric} & \textbf{Explanation} \\
 \midrule
 & \begin{minipage}{2.3cm}\includegraphics[width=0.9\linewidth]{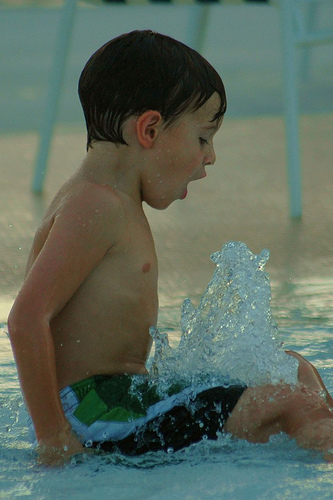}\end{minipage}\begin{minipage}{11.7cm}Reference Caption:\\``A boy sitting in water .''\\``A small boy is sitting in the water and water is splashing up through his legs .''\\``A young boy in a swimming suit sits in water .''\\``A young boy sitting on a water jet in a pool .''\\``Little boy sitting in water with a fountain coming up through his lap .''\\ \\ Candidate Caption:\\``A man and a girl are at the shoreline while the man takes a picture and the girl hods a doll''\end{minipage} \\
CLAIR & Score: 0.3 \\
& Explanation: The candidate set mentions a man and a girl, while the reference set mentions a boy. The actions described in the candidate set (taking a picture and holding a doll) are not mentioned in the reference set. The reference set also mentions water splashing up through the boy's legs, a swimming suit, a water jet, and a fountain, which are not mentioned in the candidate set. Overall, there are significant differences between the two sets of captions, indicating a low likelihood that they are describing the same image.\\
FLEUR  &  Score: 0.050262451171875 \\
& Explanation: The caption does not accurately describe the image. The image shows a young boy playing in the water, with water splashing on his face, and he is not holding a doll. The presence of a man taking a picture is also not evident in the image. The caption is not clear and does not accurately describe the scene in the image.\\
 \midrule
 & \begin{minipage}{4.7cm}\includegraphics[width=0.9\linewidth]{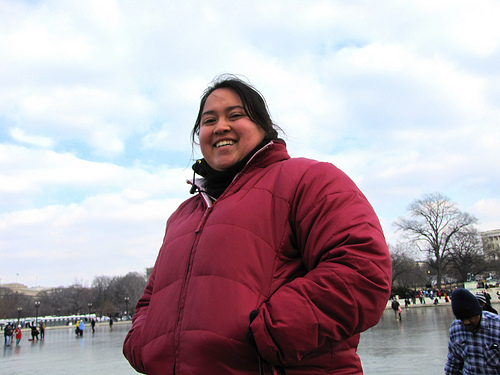}\end{minipage}\begin{minipage}{9.3cm}Reference Caption:\\``A girl in a red jacket , surrounded by people .''\\``A woman in a puffy red jacket poses for a picture at an ice skating rink .''\\``A woman in a red coat is smiling , while people in the background are walking around in winter clothing .''\\``A woman wearing a red coat smiles down at the camera .''\\``The woman in a red jacket is smiling at the camera .''\\ \\ Candidate Caption:\\``A toddler in a red jacket is balancing on top of snow .''\end{minipage}\\
CLAIR & Score: 0.7 \\
& Explanation: The candidate set and reference set share some common elements such as a person wearing a red jacket and being in a snowy environment. However, the candidate set specifically mentions a toddler while the reference set mentions a girl, woman, and people in general. Additionally, the reference set includes details about ice skating and smiling for a picture which are not mentioned in the candidate set. Therefore, while there are similarities, it is not a perfect match and the likelihood of describing the same image is moderate.\\
FLEUR  &  Score: 0.092529296875 \\
& Explanation: The caption does not accurately describe the image. The image shows a woman in a red jacket standing on a frozen lake, not a toddler in a red jacket balancing on snow. The caption is unrelated to the actual content of the image.\\
\midrule
 & \begin{minipage}{4.7cm}\includegraphics[width=0.9\linewidth]{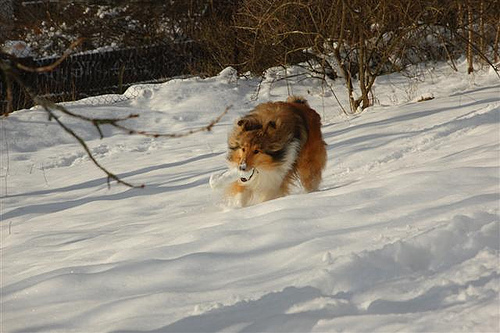}\end{minipage}\begin{minipage}{9.3cm}Reference Caption:\\``A brown dog carries an object in its mouth on a snowy hillside .''\\``A collie running through the snow .''\\``A Sheltie dog carries a white-colored toy in its mouth as it walks across the snow .''\\``Dog with a toy in the snow .''\\``The dog wades through deep snow with something in its mouth .''\\ \\ Candidate Caption:\\``A dog runs through the deep snow .''\end{minipage}\\
CLAIR & Score: 0.7 \\
& Explanation: The candidate set and reference set both mention a dog in the snow with an object in its mouth, but the candidate set does not provide as much detail about the dog's appearance or the object it is carrying. Additionally, the candidate set uses the word 'runs' while the reference set uses 'carries' and 'wades', which could indicate slightly different actions being depicted.\\
FLEUR  &  Score: 0.833984375 \\
& Explanation: The caption accurately describes the image, as it shows a dog running through the deep snow. The dog is actively moving through the snow, which is a key aspect of the scene. The caption captures the essence of the image and provides a clear description of the dog's action. Therefore, the score is 0.833984375.\\
\bottomrule
\end{tabular}
}
\caption{Additional examples of explanations for FLEUR and CLAIR.}
\label{fig:c_examples}
\end{table*}

\end{document}